\documentclass[runningheads]{llncs}

\usepackage{hyperref}
\usepackage{url}
\usepackage{xurl}
\usepackage{graphicx}
\usepackage{textcomp}
\usepackage{amsfonts}
\usepackage{xcolor}

\begin{document}

\title{Improving Results on Russian\\ Sentiment Datasets}
%
%
\author{Anton Golubev\inst{1}\orcidID{0000-0002-1883-4121} \and\\
Natalia Loukachevitch\inst{2}\orcidID{0000-0001-6553-5637}}
\authorrunning{A. Golubev \and N. Loukachevitch}
%
\institute{Bauman Moscow State Technical University, Russia \and
Lomonosov Moscow State University, Russia\\
\inst{1}\email{antongolubev5@yandex.ru}\\
\inst{2}\email{louk\_nat@mail.ru}\\}
%
\maketitle              
\begin{abstract}
In this study, we test standard neural network architectures (CNN, LSTM, BiLSTM) and recently appeared BERT architectures on previous Russian sentiment evaluation datasets. We compare two variants of Russian BERT and show that for all sentiment tasks in this study the conversational variant of Russian BERT performs better. The best results were achieved by BERT-NLI model, which treats sentiment classification tasks as a natural language inference task. On one of the datasets, this model practically achieves the human level. 

\end{abstract}

\section{Introduction}

Sentiment analysis studies are currently based on the application of deep learning approaches, which requires training and testing on specialized datasets. For English, popular sentiment analysis datasets include: Stanford Sentiment Treebank datasets SST \cite{socher2013recursive}, IMDB dataset of movie reviews \cite{maas2011learning}, Twitter sentiment datasets \cite{nakov2016semeval,rosenthal2017semeval}, and many others.
For other languages,  much less datasets have been created.  In Russian several sentiment evaluations were previously organized, including ROMIP2012-2013 and SentiRuEval2015-2016 \cite{chetviorkin2013evaluating,loukachevitch2015entity,loukachevitch2016rubtsova}, which included the preparation of annotated data on reviews (movies, books and digital cameras), news quotes, and  Twitter messages. The best results on these datasets were obtained with classical machine learning techniques such as SVM \cite{chetviorkin2013evaluating}, early neural network approaches \cite{trofimovich2016comparison}, or even engineering methods based on rules and lexicons \cite{kuznetsova2013testing}. Currently, the results achieved in the above-mentioned Russian evaluations can undoubtedly be improved.

In this study, we test standard neural network architectures (CNN, LSTM, BiLSTM) and recently appeared BERT architectures on previous Russian sentiment evaluation datasets. We compare two variants of Russian BERT \cite{devlin2018bert} and show that for all sentiment tasks in this study the conversational variant of Russian BERT performs better.  The best results were achieved by BERT-NLI model, which treats sentiment classification problem as the natural inference task. In one of the tasks this model practically achieves the human level of sentiment analysis.

The contributions of this paper are as follows:
\begin{itemize}
    \item we renew previous results on five  Russian sentiment analysis datasets using the state-of-the-art methods,
    \item we test new conversational Russian BERT model in several sentiment analysis tasks and show that it is better than previous Russian RuBERT model, 
    \item we show that the BERT model, which treats sentiment analysis as a natural language inference task achieves  the best results on all datasets under analysis.
\end{itemize}

This paper is structured as follows. In Section 2 we present sentiment analysis datasets previously created for Russian shared tasks, best methods and achieved results in previous evaluations. Section 3 describes preprocessing steps and methods applied to sentiment analysis tasks in the current study, including several BERT-based models. Section 4 presents the achieved results. In Section 5 we analyse the errors of models on difficult examples. Section 6  describes other available Russian sentiment analysis datasets and methods applied to these datasets.

\section{Datasets}
In our study we consider five Russian datasets annotated for previous Russian sentiment evaluations: news quotes of the ROMIP-2013 evaluations \cite{chetviorkin2013evaluating} and  Twitter datasets of two  SentiRuEval evaluations 2015-2016 \cite{loukachevitch2015entity,loukachevitch2016rubtsova}. Table 1 presents the datasets under evaluation, the volumes of their training and test parts, main quality measures, achieved results, and the best methods. Table 2 contains the distribution of the datasets texts  by sentiment classes.

\begin{table}
\centering
\caption{Datasets under evaluation.}\label{previuosmethods}
\begin{tabular}{|l|c|c|c|c|c|}
\hline
Dataset & Train vol. & Test vol. & Metrics & Result & Method\\
\hline
News Quotes ROMIP-2013\footnotemark[3]  & 4260 & 5500 & $F_1\ macro$ & 62.1 & Lexicons+Rules\\
SentiRuEval-2015 Telecom\footnotemark[4]  & 5000 & 5322 & $F^{+-}_1macro$ & 50.3 & SVM\\
SentiRuEval-2015 Banks\footnotemark[4]   & 5000 & 5296  & $F^{+-}_1macro$ & 36.0 & SVM\\
SentiRuEval-2016 Telecom\footnotemark[5]    & 8643 & 2247 & $F^{+-}_1macro$ & 55.9 & 2-layer GRU\\
SentiRuEval-2016 Banks\footnotemark[5]   & 9392 & 3313 & $F^{+-}_1macro$ & 55.1 & 2-layer GRU\\
\hline
\end{tabular}
\end{table}

\footnotetext[3]{\url{http://romip.ru/en/collections/sentiment-news-collection-2012.html}}
\sloppy
\footnotetext[4]{\url{https://drive.google.com/drive/folders/1bAxIDjVz_0UQn-iJwhnUwngjivS2kfM3}}
\footnotetext[5]{\url{https://drive.google.com/drive/folders/0BxlA8wH3PTUfV1F1UTBwVTJPd3c}}

\subsection{News Quotes Dataset}
For creating news quotes collection, opinions in  direct or indirect speeches were extracted from news articles \cite{chetviorkin2013evaluating}. The task was to classify quotations as neutral, positive or negative speaker comment about the topic of the quotation. It can be seen in Table 2 that class distribution in the dataset was rather balanced. The main quality measure was $F_1\ macro$.

The participants experimented with classical machine learning approaches such as Naive Bayes and SVM classifiers, but the best results were obtained by a knowledge-based approach using a large sentiment lexicon and rules: 62.1 of $F_1$ measure and 61.6 of accuracy score. This can be explained with great variety of topics and topic-related sentiment discussed in news quotes \cite{chetviorkin2013evaluating}.

\begin{table}
\centering
\caption{Class distribution by datasets (\%).}\label{distributions}
\begin{tabular}{|l|c|c|c|c|c|c|}
\hline
& \multicolumn{3}{c|}{Train sample} & \multicolumn{3}{c|}{Test sample}\\
\cline{2-7}
\raisebox{1.5ex}[0cm][0cm]{Dataset}
& Positive & Negative  & Neutral & Positive & Negative  & Neutral\\
\hline
News Quotes ROMIP-2013 & 16 & 36 & 48 & 11 & 33 & 56\\
SentiRuEval-2015 Telecom & 19 & 32 & 49 & 10 & 23 & 67\\
SentiRuEval-2015 Banks & 7 & 34 & 59 & 8 & 15 & 79\\
SentiRuEval-2016 Telecom & 15 & 29 & 56 & 10 & 46 & 44\\
SentiRuEval-2016 Banks & 8 & 18 & 74 & 10 & 22 & 68\\
\hline
\end{tabular}
\end{table}

\subsection{Twitter Datasets}
Twitter datasets were annotated for the task of reputation monitoring \cite{amigo2013overview,loukachevitch2015entity}. The goal of Twitter sentiment analysis at SentiRuEval was to find sentiment-oriented opinions or positive and negative facts about two types of organizations: banks and telecom companies. In such a way the task can be classified as targeted (entity-oriented) sentiment analysis problem. Similar evaluations were organized twice in 2015 and 2016, during which  four target-oriented datasets were annotated (Table 1). In 2016 training datasets in both domains were constructed by uniting of training and test data of the 2015 evaluation and, therefore they were much larger in size \cite{loukachevitch2016rubtsova}.

The participating systems were required to perform a three-way classification of tweets: positive, negative or neutral. It can be seen in Table 2 that neutral class was prevailing in the datasets. For this reason,  the main quality measure was $F^{+-}_1macro$ measure, which  was calculated as the average value between $F_1$ measure of the positive class and $F_1$ measure of the negative class. $F_1$ measure of the neutral class was ignored because this category is usually not interesting to know. But this does not reduce the task to the two-class prediction because erroneous labeling of neutral tweets negatively influences on $F^{+}_1$ and $F^{-}_1$. Additionally micro-average $F^{+-}_1micro$ measures were calculated for two sentiment classes \cite{loukachevitch2015entity,loukachevitch2016rubtsova}.

It can be seen in Table 1 that the results in 2016 are much higher than in 2015 for the same tasks. There can be two reasons for this. The first reason is the larger volume of the training data in 2016. The second reason is the use by the participants of more advanced methods, including neural network models and embeddings.

\section{Methods}

We compare the following groups of sentiment analysis methods on the above-described datasets.
SVM with pre-trained embeddings is a baseline for our study. We chose FastText\footnotemark[6] embeddings (dimension 300) because of its better results compared with other types of Russian embeddings such as ELMo\footnotemark[6], Word2Vec\footnotemark[7], and GloVe\footnotemark[8] in preliminary studies. To submit data to the SVM algorithm,  averaging token embeddings in a sentence was used. Grid search mechanism from scikit-learn\footnotemark[9] framework was utilized to obtain optimal hyperparameters. 

\subsection{Preprocessing}
 Since most of the data are tweets containing noise information, significant text preprocessing was implemented. The full cycle contained the following steps:
 
 \begin{itemize}
     \item lowercase cast;
     \item replacing URLs with \textit{url} token;
     \item replacing mentions with \textit{user} token;
     \item replacing hashtags with \textit{hashtag} token;
     \item replacing emails with \textit{email} token;
     \item replacing phone numbers with \textit{phone} token;
     \item replacing emoticons with appropriate tokens like \textit{sad, happy, neutral};
     \item removing all special symbols except punctuation marks;
     \item replacing any repeated more than 2 times in a row letter with 2 repetitions of that letter;
     \item lemmatization and removing stop words.
 \end{itemize}

It is worth to note that the last point was applied only in the case of SVM and classic neural networks. For BERT-based methods it did not make any difference and gave not a considerable change of about 0.01\%. 

\footnotetext[6]{\url{http://docs.deeppavlov.ai/en/master/features/pretrained\_vectors.html}}
\footnotetext[7]{\url{https://rusvectores.org/ru/models/}}
\footnotetext[8]{\url{http://www.cs.cmu.edu/~afm/projects/multilingual\_embeddings.html}}
\footnotetext[9]{\url{https://scikit-learn.org/stable/}}

\subsection{Classical neural networks}
The first  group of methods is a set of classical convolutional and LSTM neural networks. 

The architecture of the convolutional neural network considered in this paper is based on approaches \cite{cliche2017bbtwtr,zhang2015}. Input data is represented as a matrix of size $s \times d$, where $s$ is the number of tokens in the tweet and $d$ is the dimension of the embedding space. The optimal matrix height $s=50$ was chosen experimentally. If necessary, a sentence is truncated or zero-padded to the required length. 

After that several convolution operations of various sizes are applied to this matrix in parallel. A single branch of convolution involves a filtering matrix $w \in \mathbb{R}^{h \times d}$ with the size of the convolution $h$ equal the number of words it covers. 
Then output of each branch is max-pooled. This helps to extract the most important information for each convolution, regardless of feature position in the text. After all convolution operations, obtained vectors are concatenated and sent to a fully connected layer, which is then passed through the softmax layer to give the final classification probabilities. In our model we chose the number of convolution branches equal to $4$ with windows sizes $(2, 3, 4, 5)$ respectively. To reduce overfitting, dropout layers with probability of $p=0.5$ were added after max-pooling and fully connected layers. 

The main idea of LSTM recurrent networks is the introduction of a cell state of dimension $c_t \in \mathbb{R}^d$ equal to the dimension of network, running straight down the entire chain and ability of LSTM to remove or add information to the cell state using special structures called gates. This helps to avoid the exploding and vanishing gradient problems during the backpropagation training stage in recurrent neutral networks. In our work, we chose $d$ equal to the size of token embeddings. 

Besides, we used the bidirectional LSTM (BiLSTM) model, which represents two LSTMs stacked together. Two networks read the sentence from different directions and their cell states are concatenated to obtain vector of dimension $c_t \in \mathbb{R}^{2d}$. As well as in LSTM network, this vector is sent to a fully connected layer of size 40 and then passed through a softmax layer to give the final classification probabilities.

In both LSTM architectures we used dropout to reduce over-fitting by adding a dropout layer with probability of $p=0.5$ before and after the fully connected layers. For all described neural networks we used pre-trained Russian FastText embeddings with dimension of $d=300$.

\subsection{Fine-tuning BERT model}
The second group of methods is based on two pre-trained Russian BERT models and several approaches of application of BERT \cite{devlin2018bert} to the sentiment analysis task. The utilized approaches can be subdivided into single sentence classification and constructing auxiliary sentences approach \cite{utilizingbertforsa}, which converts a sentiment analysis task into a sentence-pair classification problem. It seems possible since input representation of BERT can represent both a single sentence and a pair of sentences considering them as a next sentence prediction task.

The BERT sentence-single model uses only an initial sentence as an input and represents a vanilla BERT model with an additional single linear layer with matrix $W \in \mathbb{R}^{K \times H}$ on the top.  Here $K$ denotes the number of classes and $H$ the dimension of the hidden state. For the classification task, the first word of the input sequence is identified with a unique token \textit{[CLS]}. The input representation is constructed by summing the initial token, segment, and position embeddings for any token in the sequence. Classification probabilities distribution  is calculated using the softmax function. 

The BERT sentence-pair model architecture has some differences. The input representation converts a pair of sentences in one sequence of tokens inserting special token \textit{[SEP]} between them. The classification layer is added over the final hidden state of the first token $C \in \mathbb{R}^{H}$.

For the targeted task,  there is a label for each object of sentiment analysis in a sentence so the real name of an entity was replaced by a special token. For example, the initial tweet \textit{"Sberbank is a safe place where you can keep your savings"} is converted to \textit{"MASK is a safe place where you can keep your savings"}. 

Two sentence-pair models use auxiliary sentences and based on question answering (QA) and natural language inference (NLI) tasks. The auxiliary sentences for the targeted analysis are as follows:
\begin{itemize}
    \item pair-NLI: \textit{"The sentiment polarity of MASK is"}
    \item pair-QA: \textit{"What do you think about MASK?"}
\end{itemize}
The answer is supposed to be one from the \textit{Positive, Negative, Neutral} set.

In case of the general sentiment analysis task, there is one label per sentence and no objects of sentiment analysis to mask. So we proposed to assign the token to the whole sentence. Therefore the initial sentence \textit{"56\% of Rambler Group was sold to Sberbank"}  is converted to \textit{"MASK = 56\% of Rambler Group was sold to Sberbank".} The same auxiliary sentences were  constructed for this task.

In our study, we compare two different pre-trained BERT models from DeepPavlov framework \cite{deeppavlov}:
\begin{itemize}
    \item RuBERT, Russian, cased, 12-layer, 768-hidden, 12-heads, 180M parameters, trained on the Russian part of Wikipedia and news data\footnotemark[10].
    \item Conversational RuBERT, Russian, cased, 12-layer, 768-hidden, 12-heads, 180M parameters, trained on OpenSubtitles, Dirty, Pikabu, and Social Media segment of Taiga corpus\footnotemark[10].
\end{itemize}
During the fine-tuning procedure, we set dropout probability at $0.1$, number of epochs at $5$, initial learning rate at $2e-5$, and batch size at $12$.

\footnotetext[10]{\url{http://docs.deeppavlov.ai/en/master/features/models/bert.html}}

\section{Results}

To compare different models, we calculated standard metrics such as accuracy and $F_1\ macro$. Besides, we calculated the metrics necessary for comparison with the participants of the competition: $F^{+-}_1macro$ and $F^{+-}_1micro$, which take into account only positive and negative classes. All the reported results were obtained by averaging over five runs. To distinguish two pre-trained BERT models,  special label (C) is used for Conversational RuBERT.

\begin{table}[h]
\centering
\caption{Results on News Quotes Dataset.}\label{tabromip}
\begin{tabular}{|l|c|c|c|c|}
\hline
Model & Accuracy & $F_1\ macro$ & $F^{+-}_1macro$   & $F^{+-}_1micro$\\
\hline
ROMIP-2013 \cite{chetviorkin2013evaluating} & 61.60 & 62.10 & --& --\\
SVM & 69.12 & 61.63 & 74.82 & 75.07\\
CNN & 68.57 & 60.43 & 73.51 & 74.55\\
LSTM & 73.61 & 62.31 & 77.02 & 78.20\\
BiLSTM & 74.14 & 62.78 & 77.61 & 78.94\\
BERT-single & 78.90 & 68.07 & 84.33 & 84.45\\
BERT-pair-QA & 79.06 & 68.54 & 84.33 & 84.45\\
BERT-pair-NLI & 79.68 & 69.45 & 84.96 & 85.08\\
BERT-single (C) & 79.81 &  \textbf{71.12} & 85.05 & 85.10\\
BERT-pair-QA (C) & 78.95 & 70.16 & 84.71 & 84.83\\
BERT-pair-NLI (C) &  \textbf{80.28} & 70.62 &  \textbf{85.52} &  \textbf{85.68}\\
\hline
\end{tabular}
\end{table}

\subsection {Results of News Quotes Dataset}

Table \ref{tabromip} describes results of the models on the ROMIP-2013 news quotes dataset. 
\begin{table}[h]
\centering
\caption{Results on SentiRuEval-2015 Telecom Operators Dataset.}\label{2015telecom}
\begin{tabular}{|l|c|c|c|c|}
\hline
Model & Accuracy & $F_1\ macro$ & $F^{+-}_1macro$   & $F^{+-}_1micro$\\
\hline
SentiRuEval-2015 \cite{loukachevitch2015entity} & -- & -- & 48.80 & 53.60 \\
SVM & 62.86 & 58.29 & 50.27 & 54.70\\
CNN & 60.80 & 57.52 & 49.92 & 53.23\\
LSTM & 64.46 & 58.94 & 52.10 & 56.03\\
BiLSTM & 65.54 & 59.35 & 53.01 & 56.83\\
BERT-single & 72.48 & 67.04 & 58.43 & 62.53\\
BERT-pair-QA & 74.00 & 67.83 & 58.15 & 62.92\\
BERT-pair-NLI & 74.66 & 68.24 & 59.17 & 64.13\\
BERT-single (C) & 76.55 & \textbf{69.12} & 61.34 & 66.23\\
BERT-pair-QA (C) & \textbf{76.63} & 68.54 & \textbf{63.47} & \textbf{67.51}\\
BERT-pair-NLI (C) & 76.40 & 68.83 & 63.14 & 67.45\\
Manual & -- & -- & 70.30  & 70.90\\
\hline
\end{tabular}
\end{table}
As it was mentioned before, the participants of the evaluation applied traditional machine learning methods (SVM, Naive Bayes classifier, etc.) and knowledge-based methods with lexicons and rules. The knowledge-based methods achieved the best results. This was explained by thematic diversity of news quotes,  when the test collection could contain sentiment words and expressions absent in the training collection. 
\begin{table}[h]
\centering
\caption{Results on SentiRuEval-2015 Banks Dataset.}\label{2015banks}
\begin{tabular}{|l|c|c|c|c|}
\hline
Model & Accuracy & $F_1\ macro$ & $F^{+-}_1macro$   & $F^{+-}_1micro$\\
\hline
SentiRuEval-2015 \cite{loukachevitch2015entity} & -- & -- & 36.00 & 36.60 \\
SVM & 49.23 & 43.39 & 33.08 & 36.62\\
CNN & 47.91 & 42.87 & 31.62 & 34.18\\
LSTM & 51.89 & 44.12 & 35.85 & 39.55\\
BiLSTM & 53.21 & 46.43 & 36.93 & 40.18\\
BERT-single & 83.78 & 74.57 & 57.82 & 60.64\\
BERT-pair-QA & 84.24 & 75.34 & 56.65 & 57.41\\
BERT-pair-NLI & 85.14 & 77.59 & 60.46 & 63.15\\
BERT-single (C) & 85.80 & 78.71 & 64.90 & 66.95\\
BERT-pair-QA (C) & 86.28 & 78.62 & 62.37 & 67.27\\
BERT-pair-NLI (C) & \textbf{86.88} & \textbf{79.51} & \textbf{67.44} & \textbf{70.09}\\
\hline
\end{tabular}
\end{table}
It can be seen in the current evaluation, that the task was difficult even for some models with embeddings (SVM, CNN, LSTM, BiLSTM). Among traditional neural network approaches, BiLSTM obtained the best results.

\begin{table}
\centering
\caption{Results on SentiRuEval-2016 Telecom Operators Dataset.}\label{2016telecom}
\begin{tabular}{|l|c|c|c|c|}
\hline
Model & Accuracy & $F_1\ macro$ & $F^{+-}_1macro$   & $F^{+-}_1micro$\\
\hline
SentiRuEval-2016 \cite{loukachevitch2016rubtsova} & -- & -- & 55.94& 65.69\\
SVM & 65.89 & 55.34 & 53.13 & 65.87\\
CNN & 65.28 & 54.87 & 52.62 & 64.40\\
LSTM & 66.71 & 56.74 & 56.93 & 67.18\\
BiLSTM & 67.30 & 57.11 & 57.23 & 67.93\\
BERT-single & 72.85 & 65.12 & 60.29 & 71.70\\
BERT-pair-QA & 74.24 & 66.34 & 63.86 & 73.26\\
BERT-pair-NLI & 74.51 & 67.48 & 62.81 & 73.39\\
BERT-single (C) & 75.20 & 67.89 & 64.96 & 73.91\\
BERT-pair-QA (C) & 75.27 & 68.11 & 65.91 & \textbf{74.22}\\
BERT-pair-NLI (C) & \textbf{75.71} & \textbf{68.42} & \textbf{66.07} & 74.11\\
\hline
\end{tabular}
\end{table}

The use of BERT drastically improves the results. Better results are achieved by conversational RuBERT models. The best configuration is BERT-pair-NLI, when additional \textit{MASK} token is assigned to the whole sentence and the sentence inference task was set.

\subsection{Results on Twitter Datasets}

Tables \ref{2015telecom} and \ref{2015banks} describe results of the models on two Twitter datasets of SentiRuEval-2015. The specific feature of this evaluation was a long 6 months period of time between downloading the training and test collections. In this period Ukrainian topics of tweets about telecom operators and banks led to great differences between the training and test collections.

These differences between collections showed up in very low obtained results on the bank 2015 dataset \cite{loukachevitch2015entity}. The problem  was also complicated for the current SVM+FastText, CNN, LSTM and BiLSTM models.  Only BERT-based methods could significantly improved the results. Conversational RuBERT in the NLI setting was the best method again.

It is interesting to note that one participant of the SentiRuEval-2015 uploaded manual annotation of the test Telecom dataset and obtained the results described in Table \ref{2015telecom} as Manual \cite{loukachevitch2015entity}. It can be seen that the best BERT results are very close to the manual labeling.


\begin{table}
\centering
\caption{Results on SentiRuEval-2016 Banks Dataset.}\label{2016banks}
\begin{tabular}{|l|c|c|c|c|}
\hline
Model & Accuracy & $F_1\ macro$ & $F^{+-}_1macro$   & $F^{+-}_1micro$\\
\hline
SentiRuEval-2016 \cite{loukachevitch2016rubtsova} & -- & -- &55.17& 58.81\\
SVM & 66.46 & 57.85 & 51.12 & 53.74\\
CNN & 67.15 & 58.43 & 52.06 & 54.96\\
LSTM & 70.80 & 61.17 & 57.22 & 59.71\\
BiLSTM & 71.44 & 61.86 & 58.40 & 61.06\\
BERT-single & 81.20 & 73.21 & 68.19 & 69.56\\
BERT-pair-QA & 80.35 & 72.61 & 66.61 & 68.18\\
BERT-pair-NLI & 80.91 & 72.68 & 65.62 & 67.65\\
BERT-single (C) & 80.47 & 72.59 & 66.95 & 69.46\\
BERT-pair-QA (C) & \textbf{82.28} & \textbf{74.06} & \textbf{69.53} & \textbf{71.76}\\
BERT-pair-NLI (C) & 81.28 & 73.34 & 65.82 & 68.03\\
\hline
\end{tabular}
\end{table} 

Tables \ref{2016telecom} and \ref{2016banks} describe results of the models on two Twitter datasets of SentiRuEval-2016. In contrast to previous evaluations, baseline results  of the 2016 competition (the best results achieved by participants) are better than the SVM+FastText and CNN models. This is due to the  fact that the participants applied neural network architectures with embeddings and combined the SVM method with existing Russian sentiment lexicons \cite{trofimovich2016comparison,loukachevitch2016rubtsova}.

\section{ Analysis of Difficult Examples}

The authors of previous Russian sentiment evaluations described examples, which were  difficult for most participants of the shared tasks \cite{chetviorkin2013evaluating,loukachevitch2015entity,loukachevitch2016rubtsova}. We gathered these examples and obtained the collection of 21 difficult samples. Now we can compare the performance of the models on this collection.

The difficult examples are translated from Russian and can be subdivided into several groups.

The first type of difficulties concerns the problem of the absence of a sentiment word or word with positive or negative connotations in the training collection, which was a serious problem for previous approaches.  From this group one example was again erroneously classified by all current models:
\begin{itemize}
    \item \textit{"Sberbank imposes credit cards".} (Ex.1)
\end{itemize}

The following sentence from this group was successfully processed by all models: 

\begin{itemize}
    \item \textit{"In the capital there was a daring robbery of Sberbank".} (Ex.2)
\end{itemize}

The second groups comprises  examples with complicated word combinations that include words of different sentiments and/or sentiment operators. From these examples, the following example was  problematic for all models:

\begin{itemize}
    \item
\textit{"Secretary of the Presidium of the General Council of United Russia, State Duma Deputy Chairman Sergei Neverov said on Saturday that the party is \textbf{not afraid of a split} due to the appearance 
different ideological platforms in it".} (Ex.3)
\end{itemize}

In the above-mentioned sentence there are two negative words and negation, which inverts negative sentiment to positive: \textit{"not afraid of a split".} But the following example  was processed correctly by most BERT-based models:

\begin{itemize}
    \item 
\textit{"VTB-24 \textbf{reduced losses} in the second quarter".}  (Ex.4)
\end{itemize}

The third group includes  tweets with irony. The following example was differently treated by the models: 

\begin{itemize}
    \item 
\textit{"Sberbank -- the largest network of non-working ATMs in Russia".}  (Ex.5)
\end{itemize}

The fourth group includes tweets that mention two telecom operators with different sentiment attitudes. In most cases it was difficult for models to distinguish correct sentiment towards each company.  

\begin{itemize}
    \item 
\textit{"I always said to you that the best operator is Beeline. Megaphone does not respect you".}  (Ex.6 - Beeline, Ex.7 - Megaphone)
\end{itemize}

\begin{table}[h]
\centering
\caption{Analysis of difficult examples. "Acc." means the accuracy of classification on whole collection of 21 difficult examples.}\label{difficultexamples}
\begin{tabular}{|l|c|c|c|c|c|c|c|c|c|c|c|c|c|}
\hline
Example & True & SVM & CNN & LSTM & BiLSTM & BS & BPQ & BPN & BS-C & BPQ-C & BPN-C\\
\hline
Ex.1 & \textbf{-1} & 0 & 0 & 0 & 0 & 0 & 0 & 0 & 0 & 0 & 0\\
Ex.2 & \textbf{-1} & \textbf{-1} & \textbf{-1} & \textbf{-1} & \textbf{-1} & \textbf{-1} & \textbf{-1} & \textbf{-1} & \textbf{-1} & \textbf{-1} & \textbf{-1}\\
Ex.3 & \textbf{1}& -1 & -1 & -1 & -1 & -1 & -1 & -1 & -1 & -1 & -1\\
Ex.4 & \textbf{1} & -1 & -1 & 0 & -1 & \textbf{1} & -1 & \textbf{1} & \textbf{1} & -1 & \textbf{1}\\
Ex.5 & \textbf{-1} & \textbf{-1} & \textbf{-1} & \textbf{-1} & \textbf{-1} & 0 & 0 & 0 & 0 & \textbf{-1} & \textbf{-1}\\
Ex.6 & \textbf{1} & 0 & 0 & 0 & 0 & -1 & -1 & -1 & -1 &
-1 & -1\\
Ex.7 & \textbf{-1} & 0 & 0 & \textbf{-1} & \textbf{-1} & \textbf{-1} & \textbf{-1} & \textbf{-1} & \textbf{-1} &
\textbf{-1} & \textbf{-1}\\
 
\hline
Acc.& & 0.33 &0.24 &0.48 &0.52 &0.48 &0.53 &0.62 & 0.62&0.57 &\textbf{0.71} \\
\hline
\end{tabular}
\end{table}

Table \ref{difficultexamples} describes the results of the models on difficult examples. Due to limited
space, acronyms of corresponding BERT architectures from previous tables were used. Here $-1,0,1$ denote negative, neutral and positive sentiments respectively. Correct predictions are in bold. The last row is share of correct answers for each model. The best results are achieved by BERT-pair-NLI model with pre-trained Conversational  RuBERT.

\section{Related Work}

The most latest and largest Russian sentiment dataset is RuSentiment \cite{rogers2018rusentiment}, which contains more than 30000 posts from VKontakte (VK), the most popular social network in Russia.  Each post is labeled with one of  five classes. The authors evaluated several traditional machine learning methods (logistic regression, linear SVM, Gradient Boosting) and neural networks. The best result (71.7 $F_1$ measure) was achieved by the neural network with four full-connected layers and FastText embeddings trained on VKontakte posts. In \cite{yuadaptation} the authors applied to the RuSentiment dataset multilingual BERT and RuBERT, trained on Russian text collections and obtained $F_1$ measure  87.73 by RuBERT.

Another popular dataset for Russian sentiment analysis is a tweet collection  with automatic annotations based on emoticons (RuTweetCorp) \cite{rubtsova2015constructing}. This corpus contains more than 200 thousand Twitter messages  posted  in 2013-2014 annotated as positive and   negative. 

In \cite{rubtsova2018reducing}, SVM with Word2Vec embeddings were applied the RuTweetCorp dataset. The authors of \cite{svetlov2019sentiment} tested LSTM+CNN and BiGRU models on RuSentiment and RuTweetCorp datasets.  Zvonarev and  Bilyi 
\cite{zvonarev2019comparison} compared logistic regression, XGBoost classifier and Convolutional Neural Network on RuTweetCorp and obtained the best results with CNN. 

The authors of \cite{loukachevitch2018extracting} created the RuSentRel corpus 
consisted of analytical articles devoted  to  international  relations.  The corpus is annotated with    sentiment attitudes  towards  mentioned  named  entities.  Rusnachenko et al. \cite{rusnachenko2019distant} study extraction of sentiment attitudes  using CNN and distant supervision approach on the RuSentRel corpus.

\section{Conclusion}
In this study, we tested standard neural network architectures (CNN, LSTM, BiLSTM) and recently appeared BERT models on previous Russian sentiment evaluation datasets. We applied not only vanilla BERT classification approach, but reformulation of the classification task and question-answering (QA) and natural-language inference (NLI) tasks. We also compared two variants of Russian BERT and showed that for all sentiment tasks in this study the conversational variant of Russian BERT is better. 

The best results were mostly  achieved by BERT-NLI model. In one of the tasks this model practically achieved the human level of sentiment analysis.

The source code\footnotemark[11] and all sentiment datasets\footnotemark[12] used in this work are publicly available.

\footnotetext[11]{\url{https://github.com/antongolubev5/Targeted-SA-for-Russian-Datasets}}

\footnotetext[12]{\url{https://github.com/LAIR-RCC/Russian-Sentiment-Analysis-Evaluation-Datasets}}

\section*{Acknowledgments}
The reported study was funded by RFBR according to the research project \textnumero~20-07-01059.

\bibliographystyle{splncs04}
\bibliography{sentiment.bib}

\end{document}